%% file: cpmi.tex
\documentclass[11pt]{article}
\usepackage{acl2013}
\usepackage{color}
\usepackage{times}
\usepackage{url}
\usepackage{latexsym}
\usepackage{graphicx}
\usepackage[noend]{algorithmic}
\usepackage{algorithm}
\usepackage{rotating}
\usepackage{verbatim}
\usepackage{amsmath}
\usepackage{multirow}

\newtheorem{definition}{Definition}

\newcommand{\wf}{\ensuremath{f}}
\newcommand{\wfs}{\ensuremath{\widehat{f}^s}}

\newcommand{\ignore}[1]{}

\title{Improving Pointwise Mutual Information (PMI) by Incorporating Significant Co-occurrence}

\author{Om P. Damani \\
  IIT Bombay \\
  {\tt damani@cse.iitb.ac.in} }

\date{}

\begin{document}
\maketitle

\input{tex/abstract}
\input{tex/introduction}

\input{tex/significancetest}

\input{tex/performance}



\bibliographystyle{acl}

\bibliography{bib/references}

\end{document}

%% file: tex/abstract.tex
\begin{abstract}

We design a new co-occurrence based word association
measure by incorporating the concept of significant co-occurrence in  the popular word association measure Pointwise Mutual Information (PMI). By
extensive experiments with a large number
of publicly available datasets we show that the newly introduced measure performs better than  other co-occurrence based measures and despite being resource-light, compares well with the best known resource-heavy distributional similarity and knowledge based word association measures. We investigate the source of this performance improvement and find that of the two types of significant co-occurrence - corpus-level and 
document-level,   the concept of corpus level significance combined with the use of document counts in place of word counts is responsible for all the performance gains observed. The concept of  document level significance is not helpful for PMI adaptation.

\end{abstract}

%% file: tex/introduction.tex
\section{Introduction}
\label{sec:intro}
Co-occurrence based word association measures like
PMI, LLR, and Dice are popular since they are
easy to understand and computationally efficient. They measure the strength of association between two words by comparing the word pair's corpus-level bigram frequency to some
function of the unigram frequencies of the individual words.

Recently a new measure called {\em Co-occurrence Significance Ratio (CSR)} was introduced in~\cite{csr} based on the notion of significant co-occurrence. Since CSR was found to perform better than other co-occurrence measures, in this work, our goal was to incorporate the concept of significant co-occurrence in traditional word-association measures to design new measures that may perform better than both CSR and the traditional measures.

Two different notions of significant co-occurrence are employed in CSR:
\begin{itemize}
\item {\em Corpus-level
significant co-occurrence} determines whether the ratio of observed bigram occurrences to their expected occurrences across the corpus can be explained as a pure chance phenomenon, and, 
\item {\em Document-level significant co-occurrence} determines whether a large fraction of a word-pair's occurrences within a given document have smaller spans than that under a null model where the words in the document are permuted randomly. 
\end{itemize}

\begin{table*}
\centering
{
\small
\begin{tabular}{|p{3cm}|l|p{6cm}|}   
 \hline
 & without corpus &  with corpus\\ 
 & level significance & level significance \\  \hline
word-based & PMI: $log\frac{\wf(x,y)}{f(x)*f(y)/W}$  &   cPMI: 
          $log\frac{\wf(x,y)}{f(x)*f(y)/W +  \sqrt{f(x)}*\sqrt{\ln{\delta}/(-2)}}$  
  \\ \hline
document-based & PMId: $log\frac{{d}(x,y)}{d(x)*d(y)/D}$ &  cPMId: $log\frac{{d}(x,y)}{d(x)*d(y)/D +  \sqrt{d(x)}*\sqrt{\ln{\delta}/(-2)}}$   \\ \hline
with document 
level significance & PMIz: $log\frac{Z}{d(x)*d(y)/D}$ & cPMIz: $log\frac{Z}{d(x)*d(y)/D +  \sqrt{d(x)}*\sqrt{\ln{\delta}/(-2)}}$ CSR:  $\frac{Z}{E(Z) +  \sqrt{K}*\sqrt{\ln{\delta}/(-2)}}$ \\ \hline
\end{tabular}}\\
\scriptsize
\scalebox{1.0}{%
\begin{tabular}{l l }
${f}(x,y)$ & Span-constrained ($x,y$) word pair frequency in the corpus\\
$f(x),f(y)$ & unigram frequencies of $x,y$ respectively in the corpus \\
$W$ & Total number of words in the corpus \\
${d}(x,y)$ & Total number of documents in the corpus having at-least \\
 &  one span-constrained occurrence of the word pair ($x,y$)\\
$d(x),d(y)$ & Total number of documents in the corpus containing \\
 & at least one occurrence of $x$ and $y$ respectively \\
$D$ & Total number of documents in the corpus \\
$\delta$ & a parameter varying between 0 and 1 \\
$Z$ & as per Definition~\ref{def:Z} \\
$E(Z)$ & Expected value of Z as given in Section 2.2 of~\cite{csr} \\
$K$ & Total number of documents in the corpus having at-least\\
 &  one occurrence of the word pair ($x,y$) regardless of the span\\
\end{tabular} 
}
\caption{ Definitions$^0$ of PMI, CSR, and various measures developed in this work.}
\label{tab:methods}
\end{table*}

While both these notions are employed in an integrated fashion in CSR, on analyzing CSR details, we realized that these two concepts are independent and can be applied separately to any word association measure which is a ratio of some variable's observed frequency to its expected frequency. 
We incorporate the concepts of corpus-level and 
document-level significant co-occurrence in PMI to design a new measure that performs better than both PMI and CSR, as well as other co-occurrence based word association measures. 
To incorporate document level significance, we need to use document level counts instead of word level counts (this
distinction is explained in detail in Section~\ref{sec:measures}).
To investigate whether the performance gains observed are because of the concept
of significant co-occurrence or simply because of the fact that we are using document counts instead of the word counts, we also design document count based baseline version of PMI called PMId, and several intermediate variants whose definitions are given in Table~\ref{tab:methods}. 

To our surprise, we discover that the concept of  document level significant co-occurrence does not contribute to the PMI performance improvement. Two newly designed, best-performing measures cPMId and cPMIz have almost identical performance. As the definitions\footnotetext{We consider only those word-pair occurrences where inter-word distance between $x$ and $y$ is atmost $s$, the span threshold. For a particular occurrence of $x$, we get a window of size $s$ on either side within which $y$ can occur. Strictly speaking, there should be a factor $2s$ in the denominator of the formula for PMI. Since we are only interested in the relative rankings of word-pairs, we follow the standard practice of ignoring the $2s$ factor, as its removal affects only the absolute PMI values but not the relative rankings. 
} in  Table~\ref{tab:methods} show, cPMId incorporates corpus level significance in a document count based version of PMI but does not employ the concept of document level significance, whereas cPMIz employs both corpus and document level significance. This demonstrates that the concept of corpus level significance combined with document counts is responsible for all the performance gains observed.

To summarize, we make the following contributions in this work:
\begin{itemize}
\item We incorporate the notion of significant co-occurrence in PMI to design a new measure cPMId that performs better than PMI as well as other popular co-occurrence based word-association measures on both {\em free association} and {\em semantic relatedness} tasks. In addition, despite being resource-light, cPMId performs as well as the best known distributional similarity and knowledge based measures which are resource-intensive.

\item  We investigate the source of this performance improvement and find that of the two notions of significance - corpus-level and 
document-level significant co-occurrence, the concept of  document level significant co-occurrence is not helpful for PMI adaptation. The concept of corpus level significance combined with document counts is responsible for all the performance gains observed.

\end{itemize}

\section{Related Work}\label{sec:related}
Word association measures can be divided into three broad categories: knowledge based, distributional similarity based, and co-occurrence based measures. Knowledge-based measures are based on  thesauri, semantic networks, taxonomies, or other knowledge sources~\cite{Liberman2009,wikiwalk09,wikiLinkMeasure,hughes_lexical_2007}. Distributional similarity-based measures compare two words by comparing distributional similarity of other words
around them~\cite{simRelDatasets,lsaEsslli08,bollegalaMI07}. In this work, our focus is on Co-occurrence based measures and hence we do not discuss Knowledge-based and Distributional similarity-based measures further.

Co-occurrence based measures estimate association between two words by computing some function of the words unigram and bigram frequencies. Table~\ref{tab:cooccur} contains definitions of popular co-occurrence measures. The concept of document and corpus level significance can be applied to any word association measure which is defined as the ratio of a variable's observed frequency to its expected frequency. While Chi-Square ($\chi^2$), LLR, and T-test already incorporate some notion of statistical significance, among Dice, Jaccard, and PMI, only the PMI meets this requirement. Hence our focus in this work is on designing new measures by incorporating the notion of significant co-occurrence in PMI.

\newcommand{\xydash}{ \begin{array}{c}{x' \in \{x,\neg x\}}\\ { \tiny y' \in\{y,\neg y\} }\end{array}}

\begin{table}
\centering
{
\footnotesize
\begin{tabular}{|p{2.1cm}|l|}   
 \hline
Measure & Definition\\
\hline
Chi-Square($\chi^2$) & ${\displaystyle\sum_{\xydash}} \frac{\left({f}(x',y')-E{f}(x',y')\right)^2}{E{f}(x',y')}$ \\ \hline
Dice~\cite{dice} & $\frac{2{f}(x,y)}{f(x)+f(y)}$ \\ \hline
Jaccard~\cite{jaccard} & $\frac{{f}(x,y)}{f(x)+f(y)-{f}(x,y)}$ \\ \hline
Log Likelihood Ratio(LLR)~\cite{llr} & ${{\displaystyle\sum_{\xydash}}}p(x',y')log\frac{p(x',y')}{p(x')p(y')}$  \\ \hline
Pointwise Mutual Information(PMI)~\cite{churchHanks89} & $log\frac{{f}(x,y)}{f(x)*f(y)/W}$ \\ \hline
T-test & $\frac{{f}(x,y)-E{f}(x,y)}{\sqrt{{f}(x,y)\left(1-\frac{{f}(x,y)}{W}\right)}}$ \\ \hline
\end{tabular}}\\
\scriptsize 
\scalebox{1.0}{%
\begin{tabular}{l l }
$W$ & Total number of tokens in the corpus \\
$f(x),f(y)$ & unigram frequencies of $x,y$ in the corpus \\
$p(x),p(y)$ & $f(x)/W,f(y)/W $\\
${f}(x,y)$ & Span-constrained ($x,y$) word pair frequency in corpus\\
${p}(x,y)$ & ${f}(x,y)/W $\\
\end{tabular}
}
\caption{ \small Definition of popular co-occurrence based word association measures.}
\label{tab:cooccur}
\end{table}

%% file: tex/significancetest.tex
\mathchardef\mhyphen="2D

\section{Incorporating Corpus Level Significance}\label{sec:cpmi}
In~\cite{csr}, the concept of corpus level significance was introduced by bounding the probability of observing a given corpus level phenomenon under a particular null model. In the formula for PMI, the observed frequency of a word pair's occurrences is compared with its expected frequency under a null model which assumes independent unigram occurrences. Near a given occurrence of the word $x$ in the corpus, the word $y$ can be observed with probability $f(y)/W$. Hence the expected value of $f(x,y)$ is $f(x)*f(y)/W$.  Adapting from~\cite{csr} and using Hoeffding's
Inequality, the probability of observing a given deviation between $\wf(x,y)$ and its expected value $f(x)*f(y)/W$ can be bounded. For any $t>0$:
\begin{flalign}
&P[ \wf(x,y) \geq f(x)*f(y)/W + f(x) * t ] \nonumber \\ 
&\leq \exp(-2 * f(x) * t^2) \nonumber \\
&= \delta  \nonumber \label{eq:hoeffding}
\end{flalign}
The upper-bound $\delta$ ($=\exp(-2*f(x)*t^2)$) denotes the probability of observing more than $f(x)*f(y)/W + f(x)*t$ bigram occurrences in the corpus, just by chance, under the given independent unigram occurrence null model. 
With $\delta$ as a parameter ($0<\delta<1$) and $t = \sqrt{\ln{\delta}/(-2*f(x))}$, we can define a new word association measure called {\em Corpus Level Significant PMI}(cPMI) as:
\begin{eqnarray}
cPMI(x,y) =           log\frac{\wf(x,y)}{f(x)*f(y)/W + f(x)*t} \nonumber \\
=           log\frac{\wf(x,y)}{f(x)*f(y)/W + \sqrt{f(x)}*\sqrt{\ln{\delta}/(-2)}}  \nonumber\label{cpmi} 
\end{eqnarray}
where $t = \sqrt{\ln{\delta}/(-2*f(x))}$. 

By taking the probability of observing a given deviation between $\wf(x,y)$ and its expected value $f(x)*f(y)/W$ in account,
cPMI addresses one of the main weakness of PMI of working only with probabilities and completely ignoring the absolute amount of evidence. In two scenarios where all frequency ratios (that of $f(x)$, $f(y)$, $\wf(x,y)$, and $W$) are equal, PMI values will be same while cPMI value will be higher for the case where absolute number of occurrences are higher. This can be seen easily by multiplying all of $f(x)$, $f(y)$, $\wf(x,y)$, and $W$ with some constant $n$:

{\footnotesize
\begin{flalign*}
& log\frac{n*\wf(x,y)}{n*f(x)*n*f(y)/n*W + \sqrt{n*f(x)}*\sqrt{\ln{\delta}/(-2)}} \nonumber \\
&= log\frac{\wf(x,y)}{f(x)*f(y)/W + {\bf \sqrt{1/n}}*\sqrt{f(x)}*\sqrt{\ln{\delta}/(-2)}}  \nonumber \\
&> log\frac{\wf(x,y)}{f(x)*f(y)/W + \sqrt{f(x)}*\sqrt{\ln{\delta}/(-2)}} \nonumber \\
&= cPMI(x,y) \nonumber 
\end{flalign*}
}
\section{Incorporating Document Level Significant Co-occurrence}
\label{sec:significance}
Traditional measures like PMI can be viewed as working with
a null hypothesis where each word in a document is generated completely
independently of the other words in that document. With each word,
a global unigram generation probability is associated and all documents
are assumed to be generated as per a multinomial distribution.
Such a null model generates different expected span (inter-word gap) for high frequency words vs. low frequency words. In reality, if strongly associated words co-occur in a document then they do so with low span, i.e., they occur close to each-other regardless of the underlying unigram frequencies.

\subsection{Determining Document Level Significance}\label{sec:dsig}
To correct this span bias of traditional measures, a new null model is employed in~\cite{csr}. A bag of word is associated with each document. The null model assumes that the observed document is a random permutation of the associated bag of words. Given the occurrences of a word-pair in the document, if the number of occurrences with span less than a given threshold can be explained by this null model then the word pair is assumed to be unassociated in the document. Else, some form of association is assumed.
Following definitions are introduced in~\cite{csr} to formalize this concept.
\begin{definition}[span-constrained frequency]
Let $f$ be the maximum number of {\em non-overlapped occurrences} of a word-pair $\alpha$ in a
document. 
Let $\wfs (0 \leq \wfs \leq f)$ be the maximum number of non-overlapped occurrences of
 $\alpha$ with span less than a given threshold $s$. We refer to
$\wfs$ as the {\em span-constrained frequency} of  $\alpha$ in the document.
\end{definition}\label{def:span}
For a given document  of length $\ell$ and a word-pair with $f$ occurrences in it, as we vary the span threshold $s$, the number of occurrences of the word-pair with span less than $s$, i.e. its span-constrained frequency $\wfs$ varies. For a given $s$ and the $\wfs$ resulting from it, we can ask, what is the probability that $\wfs$ out of $f$ occurrences of a word-pair in a document of length $\ell$ will have span less than $s$, if the words in the document were to be permuted randomly. If this probability is less than some threshold $\epsilon$, then we can assume that the words in the pair have some tendency of co-occurring in the document. Formally,

\begin{definition}[$\epsilon$-significant co-occurrence]
Let $\ell$ be the length of a document and let $f$ be the frequency of a word-pair $\alpha$ in it. For a given a span threshold $s$, define $\pi_s(\wfs,f,\ell)$ as the probability
under the null that $\alpha$ will appear in the document with a span-constrained frequency of {\em at least} $\wfs$.

Given a probability threshold $\epsilon$ ($0< \epsilon < 1$) and a span threshold $s$, the document is said to {\em support} the hypothesis ``$\alpha$ is an $\epsilon$-significant
word-pair within the document'' if we have $[\pi_s(\wfs,f,\ell) < \epsilon]$.
\label{def:pix}
\end{definition}

The key idea is that we should concentrate on those documents where a word pair has an $\epsilon$-significant occurrence and ignore its occurrences in non $\epsilon$-significant documents. This point is more subtle than it appears. Earlier, if the span of an occurrence was less than a threshold, it was counted, else it was ignored. In the new null model, instead of an individual occurrence, all occurrences of the word-pair in the document are considered as a single unit. Either all occurrences confirm to the null model or they do not. Of course, some occurrences will have span less than the threshold while others will have higher span, but when considering significance, all occurrences in the document are considered significant or insignificant as a unit. 
This point is discussed further in Section \ref{sec:measures}.

\subsection{$\pi_s[]$ Computation Overhead}
The detailed discussion of the computation of  $\pi_s[]$ table can be found in~\cite{csr}. For our work, it suffices to know that $\pi_s[]$ table needs to be computed only once and hence it can be done offline. We use the  $\pi_s[]$ table made publicly available\footnote{http://www.cse.iitb.ac.in/~damani/papers/EMNLP11/resources.html} by CSR researchers. The use of  $\pi_s[]$ table simply entails a memory lookup and does not increase the computation cost of a measure.

\subsection{Adapting PMI for Document Level Significance}
\label{sec:measures}

Consider the cPMI definition given earlier.
One way to adapt it for document significance is to alter the numerator such that only the span-constrained bigram occurrences in $\epsilon$-significant documents are considered in computing $\wf(x,y)$.

However, this simple adaptation is problematic. 
Consider a document with $f$ occurrences of a word-pair of which span of $\wfs$ occurrences is atmost $s$, the given span threshold. 
In the definition of cPMI, the numerator takes in account only those occurrences whose span is less than $s$, i.e., only the $\wfs$ occurrences from a document. 
As discussed earlier, the $\epsilon$-significance of a document is determined by looking at all $f$ occurrences as a whole. 
In the null model, whether a particular occurrence has span less than or greater than $s$ is not so important, what matters is that span of $\wfs$ occurrences out of $f$ is at most $s$. The word-pair is considered an $\epsilon$-significant pair within the document if the observed span of all $f$ occurrences of the pair can be explained by the null model. Hence, when adapting for $\epsilon$-significance, it is improper to count only $\wfs$ occurrences out of $f$. 

The way out of this difficulty is to count the documents and not the words. We do this adaptation in two steps. First, we replace the word counts with document counts in the definition of cPMI, giving a new measure called {\em Corpus Level Significant PMI based on Document count} (cPMId):

{\scriptsize
\begin{eqnarray}
 cPMId(x,y) &=& log\frac{{d}(x,y)}{d(x)*d(y)/D + \sqrt{d(x)}*\sqrt{\ln{\delta}/(-2)}}  \nonumber \label{cpmid}
\end{eqnarray}
}
where ${d}(x,y)$ indicates the number of documents containing at least one span constrained occurrence of $(x,y)$, and $d(x)$ and $d(y)$ indicate the number of document containing $x$ and $y$, $D$ indicates the total number of documents in the corpus, and as before, $\delta$ is a parameter varying between 0 and 1.

Having replaced the word counts with document counts, we now incorporate the concept of document level significant co-occurrence (as discussed in Section~\ref{sec:dsig}) in cPMId by replacing ${d}(x,y)$ in numerator with $Z$
which is defined as:
\begin{definition}[Z]
Let $Z$ be the number of documents that
support the hypothesis ``the given word-pair is an $\epsilon$-significant word-pair'', i.e., $Z$ is the number of documents for which $\pi_s(\wfs,f,\ell) < \epsilon$.
\end{definition}\label{def:Z}
The new measure is called {\em Document and Corpus Level Significant PMI} (cPMIz) and is defined as:

{\scriptsize
\begin{eqnarray}
 cPMIz(x,y) &=& log\frac{Z}{d(x)*d(y)/D + \sqrt{d(x)}*\sqrt{\ln{\delta}/(-2)}}  \nonumber \label{cPMIz} 
\end{eqnarray}
}

Note that cPMIz has three parameters: span threshold $s$, the corpus level significant parameter $\delta$ ($0< \delta < 1$) and the document level significant parameter $\epsilon$ ($0< \epsilon < 1$). In comparison, cPMI/cPMId have $s$ and $\delta$ as parameters while PMI has only $s$ as the parameter. The three parameters of cPMId are similar to those of CSR.

cPMIz and cPMId differ in the fact that cPMId does not incorporate the document level significance. Similarly, we can design another measure that differs from cPMIz in that it does not incorporate corpus level significance. This measure is called {\em Document Level Significant PMI} (PMIz) and is defined as:

{\small
\begin{eqnarray}
 PMIz(x,y) &=& log\frac{Z}{d(x)*d(y)/D}  \nonumber \label{PMIz}
\end{eqnarray}
}
\textbf{Baseline Measure}:
Suppose cPMIz were to do better than the PMI. One could ask whether the improvement achieved is due to the concept of significant co-occurrence or is it simply a result of the fact that we are counting documents instead of words. To answer this, we design a baseline version of PMI where we simply replace word counts with document counts. The new baseline measure is called {\em PMI based on Document count} (PMId) and is defined as:

{\scriptsize
\begin{eqnarray}
 PMId(x,y) &=& log\frac{{d}(x,y)}{d(x)*d(y)/D}  \nonumber \label{dpmi}
\end{eqnarray}
}

%% file: tex/performance.tex
\section{Performance Evaluation}
\label{section:performance}

\begin{table*}[tp] 
\small
\centering
\begin{tabular}{|l|l|l|l|l|l|l|l|l|l|l|}  
\hline   
& \begin{sideways}\parbox{15mm}{Edinburgh (83,713)}\end{sideways} & \begin{sideways}\parbox{15mm}{Florida (59,852)}\end{sideways} & \begin{sideways}\parbox{15mm}{Kent (14,086)}\end{sideways} & \begin{sideways}\parbox{15mm}{Minnesota (9,649)}\end{sideways} & \begin{sideways}\parbox{15mm}{White- Abrams (652)}\end{sideways} & \begin{sideways}\parbox{15mm}{Goldfarb- Halpern (384)}\end{sideways} & \begin{sideways}\parbox{15mm}{Wordsim (351)}\end{sideways}  & \begin{sideways}\parbox{15mm}{Esslli (272)}\end{sideways} \\  
\hline  
\hline  
PMI &0.22 &0.25 &0.35 &0.25 &0.27 &0.16 &0.69  &0.38    \\
cPMI & 0.23 & 0.28 & 0.40 & 0.29 & 0.29 & 0.17 & 0.70 & 0.46 \\ \hline
PMId &0.22 & 0.26 & 0.37 & 0.26 & 0.28 & 0.17 & 0.71 & 0.42      \\   
cPMId & \textbf{0.27} & \textbf{0.32} & \textbf{0.44} & 0.33 & \textbf{0.36} & 0.16 & \textbf{0.72} & \textbf{0.54}  \\
\hline  
PMIz &0.24 & 0.26 & 0.38 & 0.26 & 0.28 & \textbf{0.18} & 0.71 & 0.39   \\ 
cPMIz &\textbf{0.27} & \textbf{0.32} & \textbf{0.44} & \textbf{0.34} & 0.35 & \textbf{0.18} & 0.71 &0.53  \\ \hline
CSR &0.25 &0.30 &\textbf0.42 &0.31 &0.34 &0.10 &0.63  &0.43      \\
\hline
\end{tabular}

\caption{\small 5-fold cross validation comparison of rank coefficients for different measures. The number of word-pairs in each dataset is shown against its name. The best performing measures for each dataset are shown in bold. } 
\label{tab:mainresult} 
\end{table*} 

Having introduced various measures, we wish to determine whether the incorporation of corpus and document level significance improves the performance of PMI. Also, if the adapted versions perform better than PMI, what are the sources of the improvements. Is it the concept of corpus level or document level significance or both, or is the performance gain simply a result of the fact that we are counting documents instead of words? Since the newly introduced measures have multiple parameters, how sensitive is their performance to the parameter values.


To answer these questions, 
we repeat the experiments performed in~\cite{csr}, using the exact same dataset, resources, and methodology - 
the same 1.24 Gigawords Wikipedia corpus  and the same eight publicly available datasets - Edinburgh~\cite{edinburg}, Florida~\cite{florida}, Kent~\cite{kent}, Minnesota~\cite{minnesota}, White-Abrams~\cite{white-abrams}, Goldfarb-Halpern~\cite{goldfarb}, Wordsim~\cite{wordsim353}, and Esslli~\cite{esslli08}. 
Of these, Wordsim measures  {\em semantic relatedness} which encompasses relations like synonymy, meronymy, antonymy, and functional
association~\cite{budanitskyHirst}. All other datasets measure {\em free association} which  refers to
the first response given by a subject on being given a stimulus word~\cite{esslli08}.

\begin{table*}
\centering
\small
\begin{tabular}{|p{2.5cm}|l|p{2cm}|}   
 \hline
 & without corpus &  with corpus\\ 
 & level significance & level significance \\  \hline
word-based & PMI: 0.075  &   cPMI: 0.044   \\ \hline
document-based & PMId: 0.060 & cPMId: 0.004 \\ \hline
with document 
level significance & PMIz: 0.059 & cPMIz: 0.004  CSR:  0.049 \\ \hline
\end{tabular}
\caption{\small Average deviation of various measures from the best performing measure for each dataset.}
\label{tab:avgdev}
\end{table*}

\subsection{Evaluation Methodology}

Each measure is evaluated by the correlation between the ranking of word-associations produced by the measure and the gold-standard human ranking for that dataset. 
Since all methods have at least one parameter, we perform five-fold cross validation. The span parameter $s$ is varied between 5 and 50 words, and $\epsilon$ and $\delta$ are varied between 0 and 1.
Each dataset is partitioned into five folds - four for training and one for testing. 
For each association measure,  the parameter values that perform best on four training folds is used for the remaining one testing fold. The performance of a measure on a dataset is its average Spearman rank correlation over 5 runs with 5 different test folds.


\subsection{Experimental Results}
\label{sec:results}

Results of the 5-fold cross validation are shown in  Table~\ref{tab:mainresult}. 
From the results we conclude that the concept of significant co-occurrence improves the performance of PMI. The newly designed measures cPMId and cPMIz perform better than both PMI and CSR on all eight datasets.

\begin{table*}[t]
\small
\centering
\begin{tabular}{|l|l|l|l|l|l|l|l|l|}
\hline 
\begin{sideways}\parbox{15mm}{Parameters ($s$, $\delta$)}\end{sideways} & \begin{sideways}\parbox{15mm}{Edinburgh (83,713)}\end{sideways} & \begin{sideways}\parbox{15mm}{Florida (59,852)}\end{sideways} & \begin{sideways}\parbox{15mm}{Kent (14,086)}\end{sideways} & \begin{sideways}\parbox{15mm}{Minnesota (9,649)}\end{sideways} & \begin{sideways}\parbox{15mm}{White- Abrams (652)}\end{sideways} & \begin{sideways}\parbox{15mm}{Goldfarb- Halpern (384)}\end{sideways} & \begin{sideways}\parbox{15mm}{Wordsim (351)}\end{sideways}  & \begin{sideways}\parbox{15mm}{Esslli (272)}\end{sideways}  \\  
\hline
\hline  

*, 0.1 & \textbf{0.27} & \textbf{0.32} & 0.43          & \textbf{0.33} & 0.35          & 0.12          & 0.65 & \textbf{0.55} \\
*, 0.3 & \textbf{0.27} & \textbf{0.32} & \textbf{0.44} & \textbf{0.33} & \textbf{0.36} & 0.14          & 0.67 & \textbf{0.55} \\ 
*, 0.5 & \textbf{0.27} & \textbf{0.32} & 0.43          & \textbf{0.33} & \textbf{0.36} & 0.15          & 0.68 & 0.54 \\ 
*, 0.7 & \textbf{0.27} & \textbf{0.32} & 0.43          & \textbf{0.33} & \textbf{0.36} & 0.14          & 0.70 & 0.54 \\ 
*, 0.9 & \textbf{0.27} & 0.31          & 0.43          & 0.32          & 0.35          & 0.16  & \textbf{0.72} & 0.53 \\ 
\hline \hline
5w, * &  \textbf{0.27} & 0.31          & 0.43 & \textbf{0.33} & 0.35          & \textbf{0.18} & 0.66 & 0.49 \\ 
10w, * & \textbf{0.27} & \textbf{0.32} & 0.43 & \textbf{0.33} & \textbf{0.36} & \textbf{0.18} & 0.70 & 0.52 \\ 
20w, * &  \textbf{0.27} & \textbf{0.32} & 0.43 & \textbf{0.33} & \textbf{0.36} & \textbf{0.18} & 0.71 & 0.54 \\ 
30w, * &  \textbf{0.27} & \textbf{0.32} & 0.42          & 0.32       & \textbf{0.36} & \textbf{0.18} & 0.71     & 0.54 \\ 
40w, * &  \textbf{0.27} & 0.31         & 0.42           & 0.32       & 0.35          & 0.17            & 0.71          & 0.54 \\ 
50w, * &  \textbf{0.27} & 0.31          & 0.42         & 0.31        & \textbf{0.36} & 0.17          & \textbf{0.72} & 0.53 \\ 
\hline \hline

*, * &\textbf{0.27} &\textbf{0.32} &\textbf{0.44} &\textbf{0.33} &\textbf{0.36} &0.16 &\textbf{0.72}  &0.54  \\ 
20w,0.7 & \textbf{0.27} & \textbf{0.32} & 0.43 & \textbf{0.33} & \textbf{0.36} & 0.16        & 0.70 & 0.54 \\ 
50w,0.9 & \textbf{0.27} & 0.31    & 0.41     & 0.31           & 0.35            & 0.17 & \textbf{0.72} & 0.53 \\
\hline
\end{tabular}
\caption{\small 5-fold cross validation performance of cPMId for various parameter combinations. * indicates a varying parameter. } 
\label{tab:cPMIdParameters}
\end{table*}

\begin{table*}[tp] 
\small
\centering
\begin{tabular}{|l|l|l|l|l|l|l|l|l|l}  
\hline   
& \begin{sideways}\parbox{15mm}{Edinburgh (83,713)}\end{sideways} & \begin{sideways}\parbox{15mm}{Florida (59,852)}\end{sideways} & \begin{sideways}\parbox{15mm}{Kent (14,086)}\end{sideways} & \begin{sideways}\parbox{15mm}{Minnesota (9,649)}\end{sideways} & \begin{sideways}\parbox{15mm}{White- Abrams (652)}\end{sideways} & \begin{sideways}\parbox{15mm}{Goldfarb- Halpern (384)}\end{sideways} & \begin{sideways}\parbox{15mm}{Wordsim (351)}\end{sideways}  & \begin{sideways}\parbox{15mm}{Esslli (272)}\end{sideways}  \\  
\hline  
\hline  
PMI &0.22 &0.25 &0.35 &0.25 &0.27 &0.16 &0.69  &0.38    \\   
PMId &0.22 & 0.26 & 0.37 & 0.26 & 0.28 & \textbf{0.17} & 0.71 & 0.42      \\   \hline
PMI$^2$ &0.24 &0.30 &\textbf{0.43} &0.31 &0.29 &0.08 &0.62 &0.44\\
PMI$^2$d &0.23 &0.29 &0.42 &0.31 &0.30 &0.06 &0.61 &0.43\\
nPMI &0.25 &0.30 &0.41 &0.30 &0.31 &0.13 &\textbf{0.72} &0.47\\ 
nPMId &0.23 &0.26 &0.28 &0.24 &0.30 &0.15 &0.71 &0.46 \\
\hline
cPMId($\delta:0.9$) & \textbf{0.27} & \textbf{0.31}          & \textbf{0.43}          & \textbf{0.32}          & \textbf{0.35}          & 0.16  & \textbf{0.72} & \textbf{0.53} \\ 

\hline
\end{tabular}

\caption{\small 5-fold cross validation comparison of cPMId with other PMI variants. } 
\label{tab:variant} 
\end{table*}

\begin{table*}[t]
\small
\centering
\begin{tabular}{|l|l|l|l|l|l|l|l|l|}
\hline & \begin{sideways}\parbox{15mm}{Edinburgh (83,713)}\end{sideways} & \begin{sideways}\parbox{15mm}{Florida (59,852)}\end{sideways} & \begin{sideways}\parbox{15mm}{Kent (14,086)}\end{sideways} & \begin{sideways}\parbox{15mm}{Minnesota (9,649)}\end{sideways} & \begin{sideways}\parbox{15mm}{White- Abrams (652)}\end{sideways} & \begin{sideways}\parbox{15mm}{Goldfarb- Halpern (384)}\end{sideways} & \begin{sideways}\parbox{15mm}{Wordsim (351)}\end{sideways}  & \begin{sideways}\parbox{15mm}{Esslli (272)}\end{sideways}  \\  
\hline
\hline  
Dice & 0.20 & 0.27 & \textbf{0.43} & \textbf{0.32} & 0.21 & 0.09 & 0.59 & 0.36 \\ 
Jaccard & 0.20 & 0.27 & \textbf{0.43} & \textbf{0.32} & 0.21 & 0.09 & 0.59 & 0.36\\ 
$\chi^2$ & 0.24 & 0.30 & \textbf{0.43} & 0.31 & 0.29 & 0.08 & 0.62 & 0.44\\ 
LLR & 0.20 & 0.26 & 0.40 & 0.29 & 0.18 & 0.03 & 0.51 & 0.38\\ 
TTest & 0.17 & 0.23 & 0.37 & 0.26 & 0.17 & -0.02 & 0.45 & 0.33\\
\hline
cPMId($\delta:0.9$) & \textbf{0.27} & \textbf{0.31}          & \textbf{0.43}          & \textbf{0.32}          & \textbf{0.35}          & \textbf{0.16}  & \textbf{0.72} & \textbf{0.53} \\ 

\hline
\end{tabular}
\caption{\small 5-fold cross validation comparison of cPMId with other co-occurrence based measures. } 
\label{tab:comCo}
\end{table*}

\begin{table*}
\centering
\small 
\begin{tabular}{|p{10cm}| c|}      

\hline
Explicit Semantic Analysis (ESA)~\cite{Gabrilovich07computingsemantic}  & \textbf{0.74}  \\
~~~~~(reimplemented in~\cite{wikiwalk09}) & \textbf{0.71}  \\
Compact Hierarchical ESA~\cite{Liberman2009} & 0.71 \\
Hyperlink Graph~\cite{wikiLinkMeasure}  & 0.69  \\
Graph Traversal ~\cite{simRelDatasets})  & 0.66 \\
Distributional Similarity~\cite{simRelDatasets})  & 0.65  \\ 
Latent Semantic Analysis~\cite{wordsim353}  & 0.56  \\
Random Graph Walk~\cite{hughes_lexical_2007}  & 0.55   \\
Normalized Path-length (lch)~\cite{wikirelate}  & 0.55   \\
\hline
cPMId($\delta:0.9$) & \textbf{0.72}  \\ \hline
\end{tabular}
\caption{\small Comparison of cPMId with knowledge-based and distributional similarity based measures for the Wordsim dataset.}
\label{tab:prevWorkTable}
\end{table*}

\subsection{Performance Improvement Analysis}
We can infer from  Table~\ref{tab:mainresult} that the concept of corpus level significant co-occurrence and not that of document level significant co-occurrence is responsible for the PMI performance improvement. 
The Spearman rank correlation for cPMIz and cPMId are almost identical. cPMId incorporates corpus level significance in a document count based version of PMI but unlike cPMIz, it does not employ the concept of document level significance.  

To underscore this point, we also compute the difference between the correlation of each measure from the correlation of the best measure
for each data set. For each measure we can then compute the average deviation of the measure from the best performing measure across datasets.
In Table~\ref{tab:avgdev} we present these average deviations. We observe that:
\begin{itemize}
\item Average deviation reduces as we move horizontally across a row - from PMI to cPMI, from PMId to cPMId, and from PMIz to cPMIz. This shows that the incorporation of corpus level significance helps improve the performance.

\item The average deviation reduces as we move vertically from the first row to the second - from PMI to PMId, and from cPMI to cPMId. This shows that the performance gain achieved is also due to the fact that we are counting documents instead of words. 

\item Finally, the average deviation remains practically unchanged as we move vertically from the second row to the third - from PMId to PMIz, from cPMId to cPMIz. This shows that the incorporation of document level significance does not help improve the performance.
\end{itemize}

\subsection{Parameter Sensitivity Analysis}
To find out the sensitivity of cPMId performance to the parameter values, we evaluate it for different parameter combinations and present the results in Table~\ref{tab:cPMIdParameters}. To save space, we show some of the combinations only, though one can see the continuity of performance with gradually changing parameter values.

From the results we conclude that the performance of cPMId is reasonably insensitive to the actual parameter values. For a large range of parameter combinations, cPMId's performance varies marginally and most of the parameter combinations perform close to the best. 
If one does not have a training corpus then one can chose the best performing ($20w, 0.7$) as default parameter values.

As an aside, introducing extra tunable parameter occasionally reduces performance, as is the case for Goldfarb-Halpern and Esslli datasets where (*,*) is not the best performing cobination. This happens when the parameters combination that performs best on the four training fold turns out particularly bad for the testing fold.

\subsection{Comparison with other measures}
Before comparing cPMId with other measures, we note that while all co-occurrence measures being compared have span threshold $s$ as a parameter, cPMId has an extra tunable parameter $\delta$. While we would like to argue that part of power of cPMId comes from this extra tunable parameter, for an arguably fairer comparison, we would like to fix the $\delta$ value and then compare so that all methods have only one tunable parameter $s$. In Table~\ref{tab:cPMIdParameters} we find that $\delta=0.9$ performs best on the fewest number of datasets and hence we select this fixed value for comparison. However most of the conclusions that follow do not change if we were to fix some other $\delta$ value, or keep it variable.

\subsubsection{Comparison with other PMI variants}

In Section~\ref{sec:cpmi} we pointed out the PMI only works with probabilities and ignores the absolute amount of evidence.
Another side-effect of this phenomenon is that PMI over-values sparseness. All frequency ratios (that of $f(x)$, $f(y)$, and $\wf(x,y)$) being equal, bigrams composed of low frequency words get higher score than those composed of high frequency words. In particular, in case of perfect dependence, i.e. $f(x)=f(y)=\wf(x,y)$, $PMI(x,y) = log\frac{W}{f(x,y)}$. cPMId addresses this weakness by explicitly bounding the probability of observing a given deviation between $\wf(x,y)$ and its expected value $f(x)*f(y)/W$. Other researchers have addressed this issue by modifying PMI such that its upper value gets bounded. 

Since the maximum value of $\frac{\wf(x,y)}{f(x)*f(y)/W}$ is $\frac{1}{\wf(x,y)/W}$, one way to bound the former is to divide it by later. ~\cite{dailey94} defined PMI$^2$ as:
\begin{equation*}
PMI^2(x,y) =  log\frac{\frac{\wf(x,y)}{f(x)*f(y)/W}}{\frac{1}{\wf(x,y)/W}} = log\frac{\wf(x,y)^2}{f(x)*f(y)}
\end{equation*}
In~\cite{npmi}, it was noted that max. and min. value of $PMI^2$ are 0,$-\infty$, whereas one can get 1,-1 as the bounds if one normalize PMI as nPMI:
{\small
\begin{equation*}
 nPMI(x,y) = \frac{log\frac{\wf(x,y)}{f(x)*f(y)/W}}{log\frac{1}{\wf(x,y)/W}}
\end{equation*}
}
{\normalsize
In Table~\ref{tab:variant}, we compare the performance of word and document count based variants of PMI$^2$ and nPMI with PMI and cPMId. We find that 
while both nPMI and  PMI$^2$ perform better than PMI,
cPMId performs better than both variants of nPMI and PMI$^2$ on almost all datasets. 
}
\subsubsection{Comparison with other co-occurrence based measures}
In Table~\ref{tab:comCo}, we compare cPMId with other co-occurrence based measures defined in Table~\ref{tab:cooccur}. We find that cPMId performs better than all other co-occurrence based measures. Note that performance of Jaccard and Dice measure is identical to the second decimal place. This is because for our datasets $\wf(x,y) \ll f(x)$ and  $\wf(x,y) \ll f(y)$ for most word-pairs under consideration.

\subsubsection{Comparison with non co-occurrence based measures}

For completeness of comparison, we also compare the performance of cPMId with
distributional similarity and knowledge based measures discussed in Section~\ref{sec:related}. Of the datasets discussed here, these measures have only been tested on the Wordsim dataset. 
In Table~\ref{tab:prevWorkTable}, we compare the performance of cPMId with these other measures on the Wordsim dataset. We can see that cPMId compares well with the best non co-occurrence based measures.

\section{Conclusions and Future Work}
By incorporating the concept of significant co-occurrence in PMI, we get a new measure which performs better than other co-occurrence based measures. 
We investigate the source of the performance improvement and find that of the two notions of significance: corpus-level and 
document-level significant co-occurrence,   the concept of corpus level significance combined with use document counts in place of word counts is responsible for all the performance gains observed. 
We also find that the performance of the newly introduced measure cPMId is reasonably insensitive to the values of its tunable parameters.

\subsection*{Acknowledgements}
We thank Dipak Chaudhari and Shweta Ghonghe for their help with the implementation.
\newpage